\documentclass[runningheads]{llncs}
\usepackage[colorlinks=true,linkcolor=blue,citecolor=blue,urlcolor=blue]{hyperref}
\usepackage[T1]{fontenc}
\usepackage{multirow}
\usepackage{graphicx}
\usepackage{amsmath}
\newcommand{\eg}{\mbox{{\em e.g.}}}

\begin{document}
\title{Vision-Based Lane Following and Traffic Sign Recognition for Resource-Constrained Autonomous Vehicles}
%
%
\author{Md Tanjemul Islam \and
Md Rafiul Kabir
}
%
\institute{Central Michigan University, Mount Pleasant, MI, USA \\
\email{tanje2m@cmich.edu, kabir2m@cmich.edu}\\
}

\titlerunning{Vision-Based Lane Following and Traffic Sign Recognition for AVs}

\maketitle              
\sloppy

\begin{abstract}
Autonomous vehicles (AVs) rely on real-time perception systems to understand road environments and ensure safe navigation. However, implementing reliable perception algorithms on resource-constrained embedded platforms remains challenging due to limited computational resources. This paper presents a lightweight vision-based framework that integrates lane detection, lane tracking, and traffic sign recognition for embedded autonomous vehicles. A computationally efficient threshold-based lane segmentation method combined with perspective transformation and histogram-based curvature estimation is used for robust lane tracking under varying illumination conditions. A rule-based steering controller generates steering commands to maintain stable vehicle navigation. For traffic sign recognition, two lightweight convolutional neural networks (CNNs), EfficientNet-B0 and MobileNetV2, are evaluated using a custom dataset captured from the vehicle's onboard camera. Experimental results show that the system achieves real-time performance while maintaining accurate lane tracking with only 3.16\% maximum offset RMSE. EfficientNet-B0 achieves a high offline classification accuracy of 98.77\% on the test dataset, while achieving 90\% accuracy during real-time on-device deployment, outperforming MobileNetV2 in both settings. MobileNetV2, however, offers slightly faster inference and lower computational cost. These results highlight the effectiveness of lightweight vision-based perception pipelines for resource-constrained autonomous driving applications.

\end{abstract}
\section{Introduction}

In recent years, the automotive industry has focused tremendously on autonomous vehicles and intelligent transportation systems. The vision of smart cities and smart transportation infrastructures is rapidly reshaping modern mobility. The potential capabilities of these transportation technologies are wide-ranging, including improvements in safety, efficiency, comfort, and security. Ongoing advances in sensors and artificial intelligence are making these systems more reliable and efficient. This rapid technological growth is transforming not only vehicle design but also the way vehicles interact with users and their environment \cite{kabir2025digital}. One of the key requirements of autonomous driving technology is the ability to accurately perceive the surrounding environment, including the interpretation of road regulations such as traffic signs in real time \cite{marti2019review}, \cite{gruyer2017perception}. This capability enables vehicles to navigate intelligently and safely through complex urban road environments.

Among the most important perception tasks are road path detection and traffic sign recognition. For resource-constrained AVs, these tasks become even more critical, since efficient resource utilization is an important factor in achieving reliable real-time performance. Recent advances in computer vision and deep learning have significantly improved the performance of perception systems in AV. CNNs have demonstrated remarkable success in image classification tasks. However, reliable road detection and traffic sign classification can still become challenging in real-world conditions due to varying illumination, road surface variations, occlusions, complex backgrounds, motion blur, and sensor noise \cite{more2025optimizing}, \cite{gao2024traffic}. In addition, selecting appropriate algorithms and models is particularly important for autonomous systems operating on platforms with limited computational resources.

In this work, a vision-based framework is proposed that integrates lane detection, lane tracking, and traffic sign recognition for resource-constrained autonomous vehicles under varying illumination conditions. Two lightweight CNN models are used for traffic sign classification and evaluated in both offline and real-time embedded settings. It also evaluates the performance gap between offline testing and real-time deployment within a unified perception-control system.

The main contributions of this work are summarized as follows:

\begin{itemize}
\item Integration of a real-time perception and control system for resource-constrained AVs, combining lane detection, lane tracking, and traffic sign recognition on a single embedded platform.
\item A lightweight lane tracking approach using classical vision techniques, designed for stable real-time performance on resource-constrained hardware and validated under varying illumination conditions.
\item A custom traffic sign dataset collected from the onboard vehicle camera to improve real-world embedded performance and reduce domain shift.
\item Comparative evaluation of lightweight CNN architectures (EfficientNet-B0 and MobileNetV2) under both offline and real-time embedded settings under varying illumination conditions, highlighting the trade-offs.
\end{itemize}

\section{Background Study}
For AV, both road path tracking and traffic sign classification are very basic and crucial tasks. On resource-limited systems, these tasks become even more critical, as both accuracy and efficient use of resources must be carefully balanced to make the system operate smoothly.

\subsection{Lane Detection Methods}
\subsubsection{Image Processing Methods}
The core principle of image processing methods for this purpose is to extract features such as curves, edges, and boundaries. Some of the popular techniques are Canny Edge Detection and then using Hough Transformation. Those methods work by generating image gradients, which can be translated into lane boundaries, then finding straight lines that represent lane markings. Those methods are efficient and suitable for AV; however, in real-time, environmental disturbance or surrounding environmental objects could hugely affect the performance.

\subsubsection{Color and Threshold-Based Methods}
Another image processing method, which is a widely used approach, involves exploiting the color characteristics of roads and lanes. Roads' color is very distinct from the environmental elements around it, which can be isolated using color thresholding techniques. This method is computationally simple and fast, but performance can be degraded significantly under varying illumination and complex road conditions.

\subsubsection{Model-Based Detection}
Model-based techniques attempt to fit mathematical curves to detect lane curvature features. Lane curvature can be modeled using polynomial functions to represent curved roads. For improvement, estimation techniques such as RANSAC are commonly used to remove outliers and improve the accuracy of the model. The success of these model-based approaches mostly depends heavily on the accuracy of feature extraction from the input image.

\subsubsection{Deep Learning-Based }
CNNs have significantly improved the performance of road path tracking. These models efficiently extract features from lane and learn representations directly from large datasets. For example, U-Net and SegNet can classify roads, lanes, or background from the images. Deep learning-based models work accurately across a wide range of errors and environmental disturbances at the cost of high computational power.

\subsubsection{Multi-Sensor Fusion-Based}
Modern AV systems have started to use fusion-based multiple sensory approaches to be able to track the path. The main concept of fusion-based approaches is that camera-based lane detection can be integrated with LiDAR, radar, and GPS sensor data to enhance the performance of road and lane detection.

\subsection{Traffic Sign Recognition Methods}
The second most important task of the AV is to be able to classify each traffic sign efficiently and accurately.  Early classification methods mostly worked with color segmentation or shape-based detection techniques for finding specific shapes from the image, then histogram techniques were used for edge segmentation, which were finally provided to different classification techniques such as  Support Vector Machines (SVM), k-Nearest Neighbors (k-NN), or decision trees. The methods achieved reasonable performance, but in a real-time AV system, their effectiveness can be limited.

With the advancement of deep learning techniques, CNNs have become the leading approach for traffic sign classification and recognition. CNN-based models automatically learn hierarchical feature representations from an image dataset. These models have demonstrated greater and stable performance under varying illumination, scale, viewpoint conditions, motion blur, color distortion, partial blocking, and so on.  As a result, deep learning-based approaches have significantly improved the accuracy and reliability of traffic sign recognition systems in modern autonomous driving.

In this research work, the focus was on working with resource-limited AV systems. So, lightweight CNN models were only taken into consideration. Models such as MobileNet (V1, V2, and V3), ShuffleNet, SqueezeNet, and EfficientNet are such models that are specifically designed to reduce computational complexity and memory requirements while maintaining competitive classification accuracy. These architectures employ techniques such as depthwise separable convolutions, channel shuffling, parameter compression, and compound scaling to achieve efficient performance.



\begin{table}[htbp]
\centering
\caption{Comparison of lightweight CNN models for classification}
\label{tab:cnn_comparison}
\small
\begin{tabular}{l|c|c|c|p{3.9cm}} 
\hline
\textbf{Model} &  \textbf{Parameters} & \textbf{FLOPs} & \textbf{Accuracy} & \textbf{Features} \\

\hline
SqueezeNet &  $\sim$1.2M & $\sim$0.83 GFLOPs & 57.5\% & Drastically reduces parameters for lightweight applications \\
\hline
MobileNet V1 &  $\sim$4.2M & $\sim$569 MFLOPs & 70.6\% & Uses depthwise separable convolutions to reduce computation \\
\hline
ShuffleNet &  $\sim$1.4M & $\sim$140 MFLOPs & 67.4\% & Employs group convolution and channel shuffle for efficiency \\
\hline
MobileNet V2 &  $\sim$3.4M & $\sim$300 MFLOPs & 72.0\% & Introduces inverted residual blocks and linear bottlenecks \\
\hline
EfficientNet-B0 &  $\sim$5.3M & $\sim$390 MFLOPs & 77.1\% & Uses compound scaling of depth, width, and resolution for better accuracy \\
\hline
MobileNet V3 &  $\sim$5.4M & $\sim$219 MFLOPs & 75.2\% & Combines Neural Architecture Search \& attention modules \\
\hline
\end{tabular}
\end{table}

Lightweight CNN models have been proposed for embedded vision tasks \cite{Iandola2016SqueezeNetAA,howard2017mobilenets,zhang2018shufflenet,sandler2018mobilenetv2,tan2019efficientnet}, which are compared in the Table \ref{tab:cnn_comparison}. 
MobileNet V2 and EfficientNet-B0 were selected for further evaluation in this work because they have a kind of balance between improved classification accuracy and smaller model size.

\section{Related Work}
Extensive research has been conducted in this domain, evolving from traditional computer vision and image processing techniques to advanced approaches that integrate neural networks with machine learning classifiers. Several studies have explored vision-based lane detection using camera systems. For instance, a lane detection algorithm based on a single RGB camera was proposed to identify the safe drivable region in front of a vehicle by applying color-space analysis, Canny edge detection, and the Hough transform to extract lane boundaries, achieving an accuracy of approximately 95.75\% in simulation environments \cite{muthalagu2020lane}. To improve robustness under challenging driving conditions such as low visibility, occlusion, and poor lighting, another study introduced a comprehensive intensity threshold range to enhance the traditional Canny edge detector, enabling more reliable detection of faint or obscured lane markings~\cite{sultana2023vision}. Similarly, a perspective-based lane detection approach utilized region-of-interest selection, perspective warping, and color filtering in LUV and LAB color spaces to isolate lane pixels, followed by polynomial regression to estimate lane curves~\cite{sapkal2023lane}. Furthermore, another method improved lane detection accuracy by processing yellow and white lane markings separately, extracting yellow lanes in the HSV color space and white lanes in grayscale space, and applying a sliding window polynomial fitting technique to better detect curved lane boundaries compared to traditional Hough-transform-based methods~\cite{nugraha2017towards}.

CNNs automatically learn hierarchical feature representations, making them more robust than traditional approaches. Several studies have explored CNN-based solutions for traffic sign detection and classification in autonomous driving systems. For example, a deep learning-based traffic sign detection model was proposed to achieve real-time performance and high accuracy using a custom dataset containing diverse traffic sign types and challenging driving scenarios~\cite{yu2024autonomous}. To further reduce computational cost, another study introduced a CNN-based traffic sign recognition method that applies non-uniform adaptive approximation, allowing different image regions to be processed with varying levels of computational precision based on their importance~\cite{omidian2025adapts}. Another study investigated a neural network–based traffic sign recognition (TSR) system under different environmental conditions and image resolutions. The experiments evaluated recognition performance across multiple weather scenarios and resolutions. The results showed that the trained network could achieve recognition accuracy above 96\% ~\cite{tiron2019neural}. Furthermore, a platform-independent CNN-based traffic sign recognition system was developed to detect multiple traffic sign shapes, including circular, triangular, and rectangular signs, trained on 43 classes from the GTSRB dataset, demonstrating suitability for real-time embedded applications~\cite{farag2018recognition}.

\section{Implementation And Setup}
\subsection{Environment Setup}
\begin{figure}
\centering
    \includegraphics[width=.9\columnwidth]{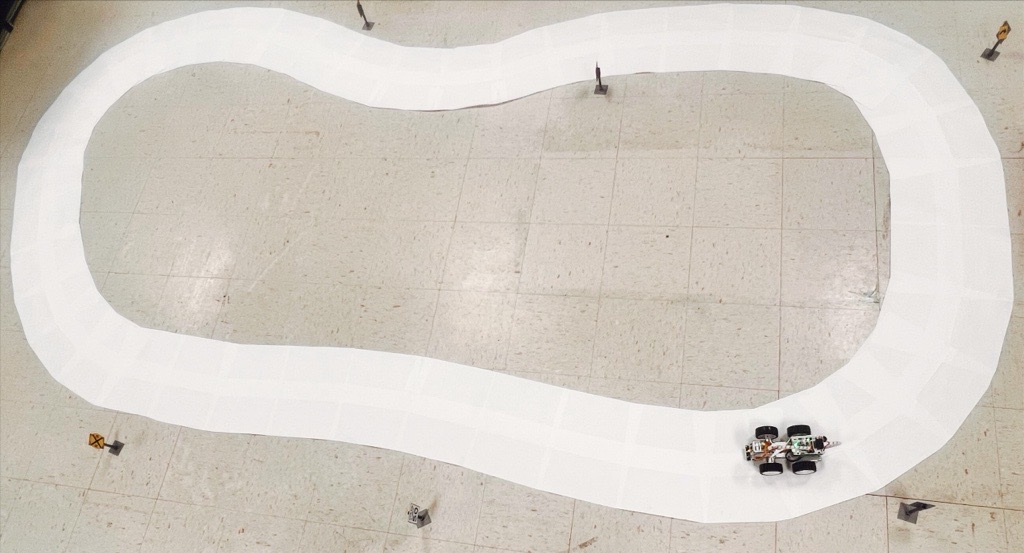}
  \caption {Pi-AV real-time deployment environment}
  	\label{fig.1}
\end{figure}

\begin{figure}[t]
\centering
    \includegraphics[width=.75\columnwidth]{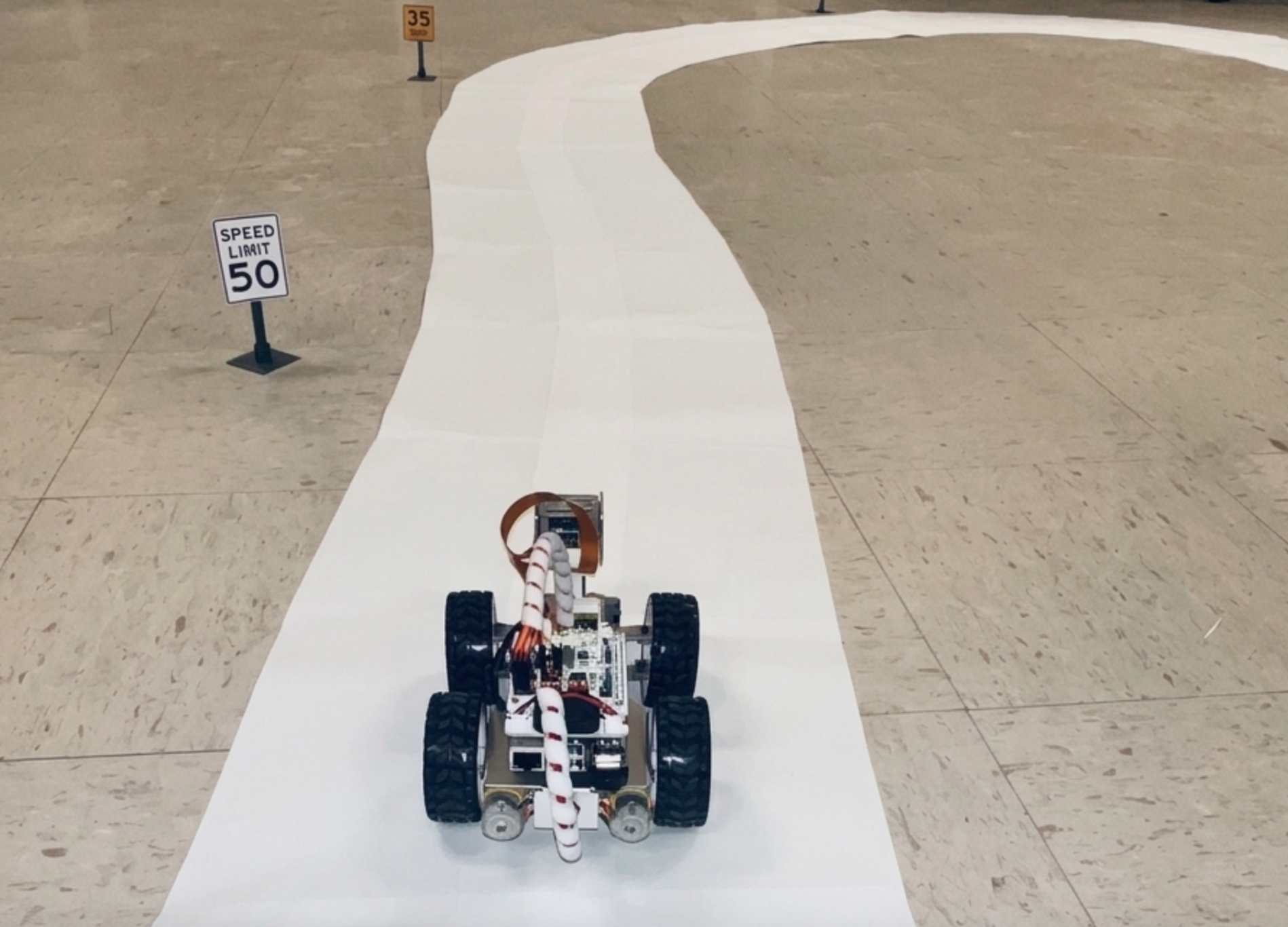}
  \caption {Deployed Pi-AV performing classification and following the track}
  	\label{fig.2}
\end{figure}

In the laboratory environment, a closed-loop track was constructed to evaluate the AV system. A small-scale AV testbed based on a Raspberry Pi platform, hereafter referred to as the Pi-AV (Pi-based Autonomous Vehicle Testbed), was deployed on a closed-loop track to evaluate the proposed perception and control framework, shown as Fig.~\ref{fig.1} and \ref{fig.2}. The Pi-AV was tasked with continuously detecting and tracking the path. Multiple traffic signs were positioned along the route to simulate real-world conditions. To assess the system’s robustness, tests were conducted under varying illumination conditions, which work as environmental perturbations for this study. During navigation, the system concurrently performed 
path tracking and traffic sign classification using a single camera as the primary input sensor.

\subsection{Lane Detection and Tracking}
For resource-constrained Pi-AV, it is important to detect the track around the surroundings efficiently with minimal computation. These robotic systems have severe computational complexity, minimal hardware requirements and low tolerance for performance. Highly computational image processing or deep learning methods can easily take away most of the computational resources. To work around this, a threshold-based segmentation method was used. The algorithm is computationally very efficient, lightweight, and still very reliable.

\begin{figure}
\centering

\begin{minipage}{0.32\textwidth}
\centering
\includegraphics[width=\textwidth]{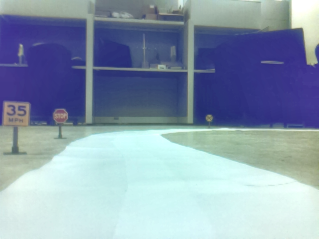}
\\ (a)
\end{minipage}
\hfill
\begin{minipage}{0.32\textwidth}
\centering
\includegraphics[width=\textwidth]{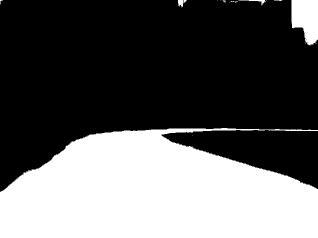}
\\ (b)
\end{minipage}
\hfill
\begin{minipage}{0.32\textwidth}
\centering
\includegraphics[width=\textwidth]{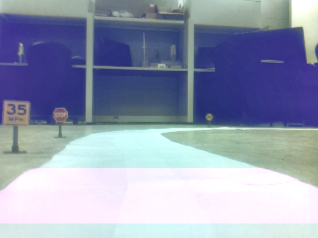}
\\ (c)
\end{minipage}

\caption{Lane curvature detection method: (a) Original Pi-AV vision, (b) Extracted path, (c) Detecting curvature of the lane.}
\label{fig.3}

\end{figure}

A color and intensity threshold is applied to the image frame, which isolates the lanes from the surrounding environment, making it easier to remove everything from the image except the track. This thresholding technique leaves a binary representation of the track, shown as Fig.~\ref{fig.3}. For simplifying the track geometry, a perspective transformation was then applied to the frame. This is often referred to as the “Bird’s-eye view”. This effectively reduces geometric distortion by aligning lane boundaries vertically. The computational part of the algorithm is the horizontal pixel intensity histogram over the lower region of the warped image. Peaks in the histogram correspond to lane boundaries, allowing a good estimation of the lane center. The difference between the detected lane center and the image midpoint is then calculated, which is basically the deviation of the Pi-AV from the lane. This is the source signal of our steering control. A rule-based steering control method was developed that converts lane center deviation into a bounded steering angle using a predefined linear relation. A smoothing mechanism is applied to ensure stable real-time motion. The resulting control signal is sent to the steering actuator for real-time lane following.

\subsubsection{Finding Threshold Value}
The accuracy of this algorithm depends highly on threshold settings. Captured RGB frames are processed to the HSV color space. Based on lower and upper thresholding values, the frames are masked to extract the track.

\begin{equation}
\mathrm{Lower} = \bigl(H_{\mathrm{low}}, S_{\mathrm{low}}, V_{\mathrm{low}}\bigr), \quad
\mathrm{Upper} = \bigl(H_{\mathrm{high}}, S_{\mathrm{high}}, V_{\mathrm{high}}\bigr)
\end{equation}

\begin{equation}
\mathrm{mask}(p) =
\displaystyle
\left\{
\begin{array}{ll}
1, & \parbox[t]{0.7\textwidth}{if $H_{\mathrm{low}} \le H(p) \le H_{\mathrm{high}},\; 
S_{\mathrm{low}} \le S(p) \le S_{\mathrm{high}},\;
V_{\mathrm{low}} \le V(p) \le V_{\mathrm{high}}$} \\
0, & \text{otherwise}
\end{array}
\right.
\end{equation}

The threshold values were empirically determined through trial-and-error for each illumination condition. The selection process involved calibrating the thresholds based on the average ambient lighting and manually tuning the upper and lower bounds to effectively isolate the track while minimizing background noise. This ensures that appropriate threshold values are selected according to luminosity conditions. The chosen thresholds are illumination-dependent and remain fixed during each experimental run.
\subsubsection{Color Space Transformation}
HSV color space provides greater flexibility in isolating and detecting chrominance components within image frames compared to the RGB space. Because of this, the frame is converted from RGB to HSV color segmentation. 
\subsubsection{Perspective Transformation}
Detection accuracy can be further improved by using a classical perspective transformation slightly. Each frame is then transferred with this perspective change to get “Bird’s eye view”, which allows to reduce the effects of perspective distortion. The transformed coordinates of a point (x,y)  are computed as:
\begin{equation}
x' = \frac{h_{11}x + h_{12}y + h_{13}}{h_{31}x + h_{32}y + h_{33}}, \quad
y' = \frac{h_{21}x + h_{22}y + h_{23}}{h_{31}x + h_{32}y + h_{33}}
\end{equation}
Where hij are elements of the 3×3 homography matrix  H.

\subsubsection{Lane Curvature Calculation}
This is the most important part, which calculates how much the curvature of the track is ahead of the vehicle that has to be proportionally offset by generating appropriate steering command signals, which eventually reduces the displacement of the vehicle from the center line of the track and keeps the vehicle on the track.  For the calculation of the curvature, a specific region of interest (ROI) is selected that is very close to the camera of the vehicle and covers some track distance in front of it. An image histogram is applied to detect the curvature from the ROI area of the image.
\subsubsection{Rule-based Steering Controller }
\begin{figure}[t]
\centering
    \includegraphics[width=.9\columnwidth]{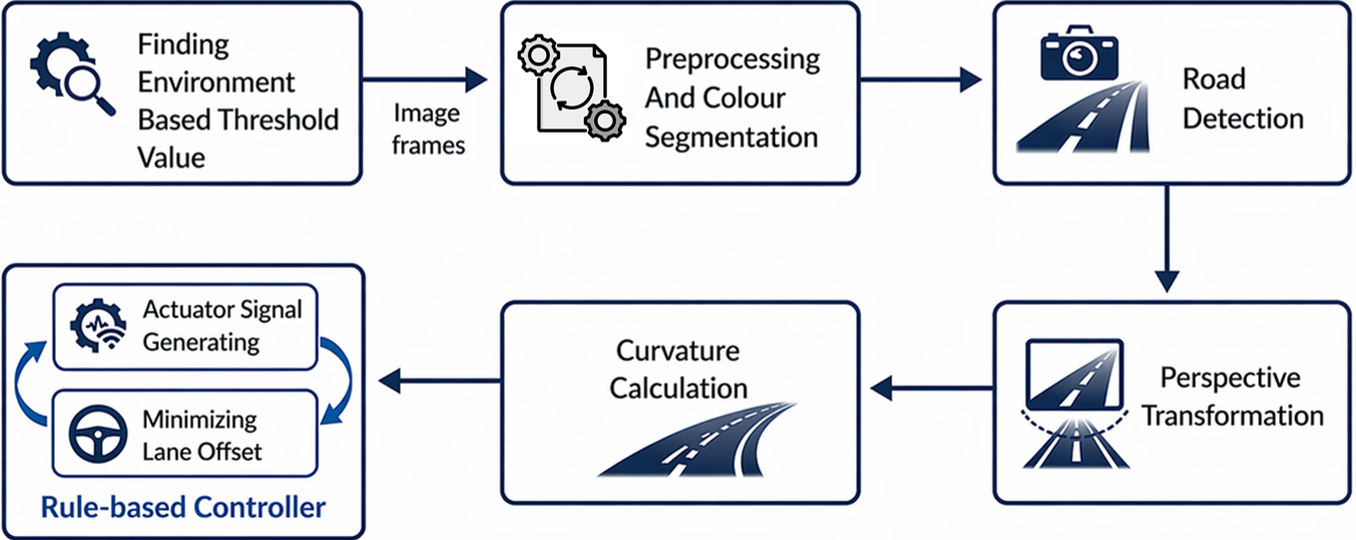}
  \caption {Path detecting and tracking block diagram}
  	\label{4}
\end{figure}

Based on the curvature calculation, the controller simultaneously provides a steering control signal that is fed to the steering servo. This enables precise vehicle maneuvers and accurate path following. Fig.~\ref{4} shows details of the overall path tracking method of the Pi-AV.

\subsection{Traffic Sign Classification}

\subsubsection{Classification with EfficientNetB0}

EfficientNetB0 belongs to the EfficientNet family. The key feature of this model is that, unlike conventional CNN architectures, which arbitrarily increase network depth or width, EfficientNet applies a principled scaling strategy that improves model performance while maintaining computational efficiency. This is called a “Compound Scaling Method”, which systematically scales network depth, width, and input resolution. The EfficientNet-B0 model is selected because of its computational efficiency, accuracy, and inference speed, which are critical factors for resource-constrained systems. 
Another feature of  EfficientNetB0 is that it uses input images of resolution 224 × 224, which is divisible by common hardware processing units (e.g., multiples of 8 or 16). This design helps to reduce unnecessary zero-padding operations and improve computational efficiency during convolution operations. 

\begin{figure}[t]
\centering
    \includegraphics[width=.7\columnwidth]{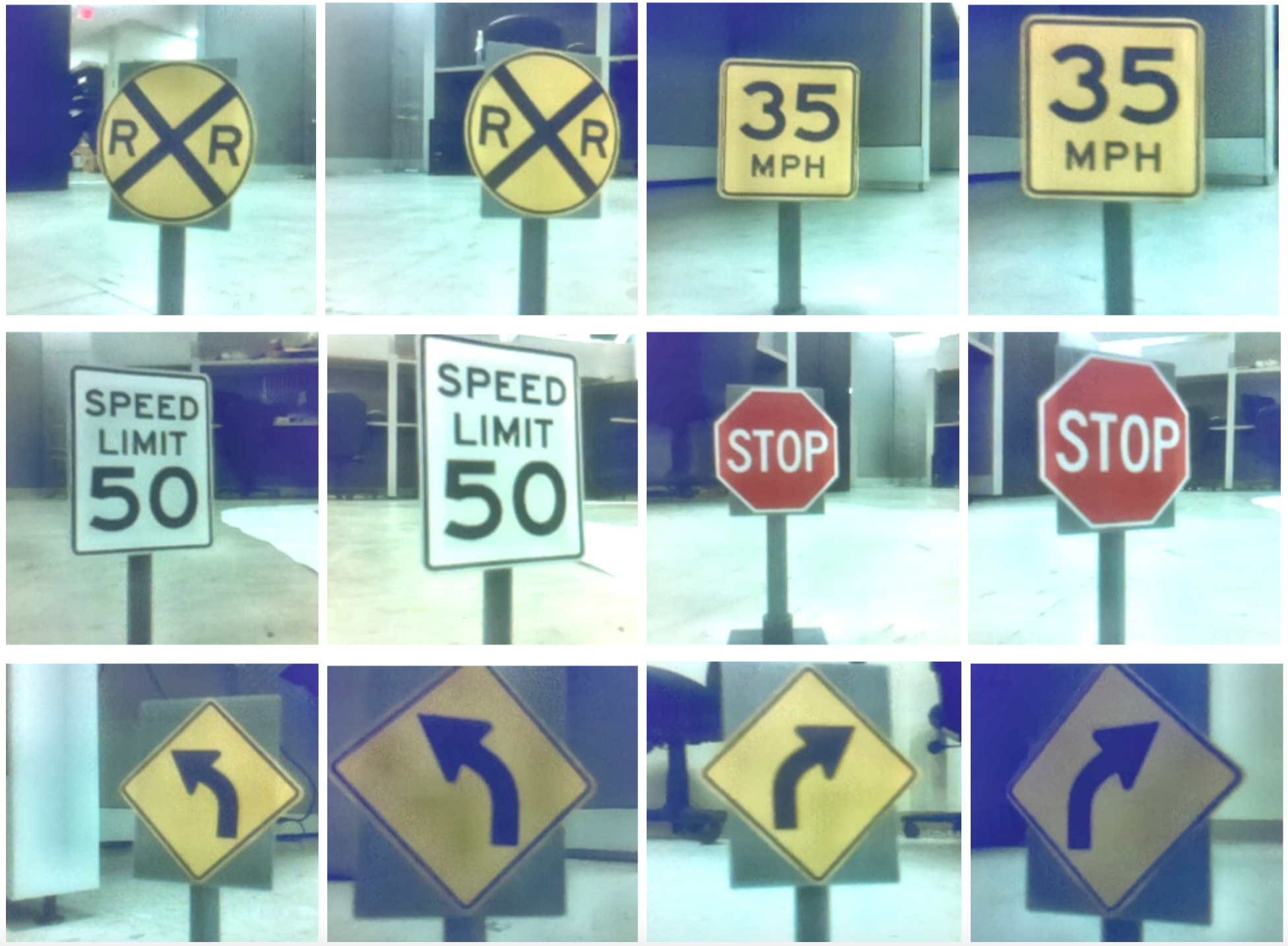}
  \caption {Sample images from training dataset}
  	\label{5}
\end{figure}

To optimize the model better, a custom traffic sign dataset was created for training the model. Public traffic sign datasets often focus on large-scale road environments and sometimes are tightly cropped. By collecting and labeling images directly from the Pi-AV camera system, the dataset captures realistic environmental variations such as viewing angles, varying illumination, motion blur, camera perspective, background clutter, and background noise specific to the robot's real-world operating environment. Some sample images from the dataset are shown in Fig.~\ref{5}. There are a total of seven classes, six are for six different traffic signs, and another is the default "None" class. Each class has 2270 images, which were taken from the original camera of the Pi-AV while it traveled through the track. The dataset was divided into train, test, and validation sets in 70\%, 15\%, and 15\%, respectively, for each class.

\subsubsection{Classification with MobileNet-V2}
MobileNetV2 was also considered for traffic sign classification due to its suitability and performance in embedded systems and robotics. MobileNetV2 is an improvement over earlier mobile-oriented CNNs. The model was specifically designed to achieve high performance while maintaining low computational memory requirements, making it appropriate for deployment on resource-constrained platforms.
The core architectural benefits of MobileNetV2 are the use of inverted residual blocks and linear bottleneck layers. Traditionally, residual networks (\eg, ResNet) use wide layers internally and narrow layers at the input and output. MobileNetV2 inverts this structure, resulting in reducing the number of parameters and floating-point operations required during inference while preserving important feature representations. Another important aspect is that MobileNetV2 employs depthwise separable convolutions, splitting convolution operations into depthwise and pointwise stages. This significantly reduces computational complexity compared to conventional CNNs while maintaining competitive accuracy. Overall, MobileNetV2's efficient architecture and lower computational demand lead to faster inference time, allowing the Pi-AV to process visual information and respond in near real-time. 

To be able to make a fair comparison between the performance of the EfficientNet-B0 model and the MobileNet-V2 model, the exact same dataset and splitting were used for training, validating, and testing both models. The training strategy was kept consistent. Exact same configuration was used for image size, data transformations, batch size, optimizer, warm-up strategy, learning rate schedule, total number of training epochs, early stopping criteria, and loss function.

\section{Result And Discussions}
In the laboratory setup, Pi-AV was deployed to operate autonomously while performing real-time object classification, while following the track. Separate experiments were conducted using the EfficientNet-B0 and MobileNetV2 models to evaluate their performance, efficiency, and relative strengths in different luminosity.
\subsection{Lane Tracking Performance}

The Pi-AV was allowed to run on the road for 5 minutes without any disturbance in two different environmental luminosity. The performance log was saved for further analysis.



\begin{table}[htbp]
\centering
\caption{Performance evaluation under different lighting conditions}
\label{tab:performance}
\setlength{\tabcolsep}{4pt}
\begin{tabular}{c| c c c}
\hline
\textbf{Avg. Luminosity} & \textbf{Avg. Inference} & \textbf{Avg. FPS} & \textbf{Avg. Lane Offset} \\
\textbf{(Lux)} & \textbf{Time (ms)} &  & \textbf{(Pixels)} \\
\hline
282.82 & $27.96 \pm 2.37$ & $35.83 \pm 1.24$ & $9.04 \pm 18.07$ \\
487.90 & $33.24 \pm 5.98$ & $30.67 \pm 3.91$ & $8.30 \pm 17.64$ \\
\hline
\end{tabular}

\vspace{6pt}

\begin{tabular}{c| c c c}
\hline
\textbf{Avg. Luminosity} & \textbf{Lane Offset} & \textbf{Normalized} & \textbf{Curvature Steering} \\
\textbf{(Lux)} & \textbf{RMSE (Pixels)} & \textbf{RMSE (\%)} & \textbf{Correlation} \\
\hline
282.82 & 20.2 & 3.16 & 0.996 \\
487.90 & 19.5 & 3.05 & 0.994 \\
\hline
\end{tabular}
\end{table}

The experimental results are shown in Table \ref{tab:performance}. It demonstrates that the system maintains real-time processing capability, achieving an average inference time of 27.96 ms and 33.24 ms at 282.82 lux and 487.90 lux, respectively. Overall, the system exhibits robust performance across varying illumination levels. Correspondingly, the system operates at an average frame rate of 35.83 FPS and 30.67 FPS, indicating that the algorithm is capable of real-time operation suitable for autonomous driving applications. In terms of tracking accuracy, the average lane offset remains relatively small for both lighting conditions, with values of 9.04 pixels and 8.30 pixels, respectively. The lane offset RMSE values of 20.2 pixels and 19.5 pixels further confirm the stability of the detection algorithm. Additionally, the normalized RMSE values remain around 3\%, suggesting that the proposed method maintains consistent accuracy despite variations in illumination.

\begin{figure}[htbp]
\centering

\begin{minipage}{0.49\textwidth}
\centering
\includegraphics[width=\textwidth]{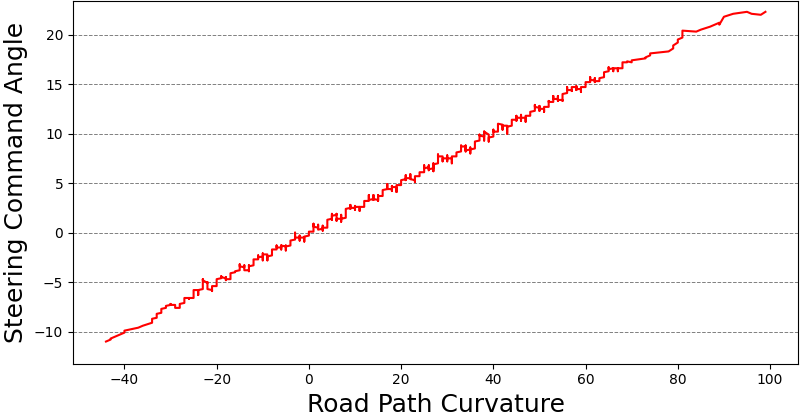}
\\ (a)
\end{minipage}
\hfill
\begin{minipage}{0.49\textwidth}
\centering
\includegraphics[width=\textwidth]{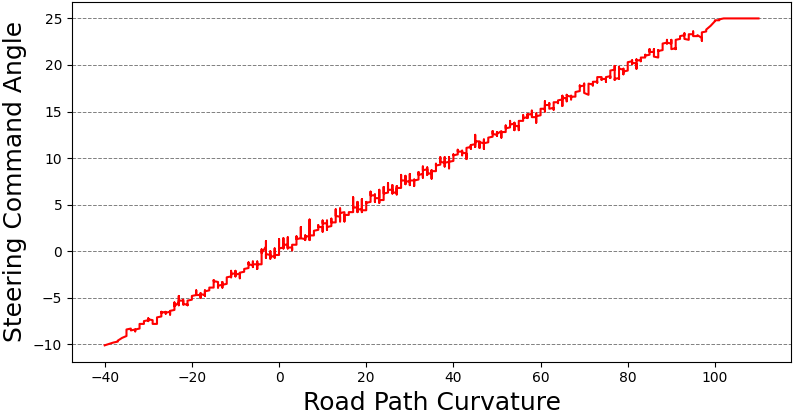}
\\ (b)
\end{minipage}

\caption{Path curvature and steering angle relationship: (a) In 282.82 lux condition and (b)  In 487.90  lux condition}
\label{6}

\end{figure}

Furthermore, the curvature–steering correlation values of 0.996 and 0.994 indicate a strong relationship between the detected track curvature and the steering control signal generated by the controller. This high correlation demonstrates that the controller responds accurately to the geometry of the detected track, enabling smooth and reliable vehicle navigation. Fig.~\ref{6} illustrates the relationship between path curvature and the steering command angle under two different illumination conditions. In both cases, a strong linear correlation can be observed, indicating that the controller generates steering commands proportional to the detected track curvature.

Overall, the experimental results confirm that the proposed system provides robust and reliable performance under varying lighting conditions while maintaining real-time processing capability.

The proposed method was compared with a baseline approach where steering angles were directly mapped from lane deviation without smoothing. The baseline method resulted in noticeable steering jitter and higher tracking error, whereas the proposed method achieved smoother and more stable lane tracking with reduced error (RMSE reduced to 3.16\% at maximum in different settings as mentioned in the Table~\ref{tab:performance} ).

\subsection{Traffic Sign Recognition Performance}
Separate experimentation with the EfficientNet-B0 and MobileNetV2 models was carried out to find their efficiency and strengths.

\subsubsection{Classification Model Test Result Comparison}

\begin{table}[htbp]
\centering
\caption{Test performance comparison of lightweight CNN models for traffic sign classification}
\label{tab:test}
\small
\setlength{\tabcolsep}{3pt}
\begin{tabular}{l|cccc}
\hline
\textbf{Model} & \textbf{Avg. Accuracy} & \textbf{Avg. Macro} & \textbf{Avg. Params} & \textbf{Avg. Inference} \\
\textbf{} & \textbf{(\%)} & \textbf{F1} & \textbf{(M)} & \textbf{(ms)} \\
\hline
EfficientNet-B0 & $98.77 \pm 0.10$ & $0.9878 \pm 0.0011$ & 4.02 & $8.63 \pm 0.007$ \\

MobileNetV2 & $95.91 \pm 1.67$ & $0.958 \pm 0.017$ & 2.23 & $7.46 \pm 0.09$ \\
\hline
\end{tabular}
\end{table}

Training was conducted for the EfficientNet-B0 model. Some randomness in data augmentations, dataset shuffling, classifier weight initialization, and dropout layers was added to conduct three iterations of training. The same steps were followed for the MobileNetV2 model training.
After training, testing was performed with the same test dataset for all the iterations. For better resemblance, very small disturbances were added, such as motion blur, color change, and sensor noise to the images.
The result is presented in Table~\ref{tab:test}. It can be observed that EfficientNet-B0 has a higher average accuracy (98.77\%) and a higher macro F1 (0.9878) compared to MobileNetV2 (95.91\% accuracy, 0.958 macro F1). The higher macro F1 score for EfficientNet-B0 indicates that it performs better across all classes. Also, EfficientNet-B0’s low standard deviation suggests it is more stable and consistent in predictions.
EfficientNet-B0 has 4.02M parameters, almost double MobileNetV2 (2.23M). This means EfficientNet-B0 is larger and more complex, which usually contributes to its higher accuracy but at the cost of more resource consumption.
MobileNetV2 is slightly faster (7.46 ms) than EfficientNet-B0 (8.63 ms). The difference (1.17 ms) is relatively small; it could be an issue based on applications, such as hardware constraints or real-time decisions. If the goal is maximum accuracy and consistent performance, EfficientNet-B0 is the better choice. If the goal is speed and minimal model size for deployment on devices with limited memory or computation, MobileNetV2 is more suitable.

\subsubsection{Real Time Performance Comparison}


\begin{table*}[htbp]
\centering
\caption{Real-time performance comparison of lightweight CNN models under different lighting conditions}
\label{tab:4}
\small
\setlength{\tabcolsep}{4pt}
\begin{tabular}{lc|ccc}
\hline
\textbf{Model} & \textbf{Avg. Luminosity} & \textbf{Accuracy} & \textbf{Macro F1} & \textbf{Avg. Inference}\\
\textbf{} & \textbf{(Lux)} & \textbf{(\%)} & \textbf{F1} & \textbf{(ms)}\\
\hline
\multirow{2}{*}{EfficientNet-B0} & 282.82 & 90.00 & 0.90 & 221.01\\
 & 487.90 & 85.70 & 0.87 & 220.22\\
\hline
\multirow{2}{*}{MobileNetV2} & 282.82 & 85.71 & 0.85 & 193.81\\
 & 487.90 & 82.86 & 0.83 & 192.51\\
\hline
\end{tabular}

\vspace{6pt}

\begin{tabular}{lc|ccc}
\hline
\textbf{Model}  & \textbf{Avg. Luminosity} & \textbf{Avg. FPS} & \textbf{Avg. Memory} & \textbf{Avg. CPU} \\
\textbf{}  & \textbf{(Lux)} & \textbf{Avg. FPS} & \textbf{Usage (MB)} & \textbf{(\%)} \\
\hline
\multirow{2}{*}{EfficientNet-B0} & 282.82  & 4.58 & 564.73 & 66.35 \\
 & 487.90  & 4.60 & 565.37 & 61.69 \\
\hline
\multirow{2}{*}{MobileNetV2} & 282.82  & 5.23 & 555.78 & 65.65 \\
 & 487.90  & 5.29 & 555.27 & 60.39 \\
\hline
\end{tabular}

\end{table*}

The models were compared using classification accuracy, Macro F1-score, inference time, frames per second (FPS), CPU usage, and memory utilization. A confusion matrix was built for each model. The evaluation results are summarized in Table~\ref{tab:4}. At an average 282.82 lux luminosity condition, EfficientNet-B0 achieved an overall accuracy of 90\% and a Macro F1-score of 0.90, outperforming MobileNetV2, which achieved 85.71\% accuracy and a Macro F1-score of 0.85.
EfficientNet-B0’s higher accuracy comes at the cost of a little bit slower inference, with an average frame rate of 4.58 FPS, compared to 5.23 FPS for MobileNetV2. The additional computational load is also reflected in slightly higher CPU and memory utilization. At an average luminance of 487.90 lux, both models' accuracies are slightly reduced. But EfficientNet-B0 still holds higher accuracy than the MobileNetV2 model; on the other hand, MobileNet has higher fps than EfficientNet. The relationship between the two models holds the same in different luminosity conditions. So the trade-off between model accuracy and real-time performance on resource-constrained embedded platforms can easily be noticed.
For real-time Pi-AV, MobileNetV2 offers a better balance between responsiveness and computational efficiency, enabling smoother operation without substantial loss of accuracy. On the other hand, EfficientNet-B0 is more precise at the cost of more hardware and time resources. If accuracy is a higher priority, then some hardware acceleration or model optimization techniques might be used to achieve comparable real-time performance.

%
%

%
\section{Conclusion}
The experimental results demonstrate that the proposed system reliably performs path detection, path tracking, and real-time traffic-sign classification using a single-camera input. The observed relationship between track curvature and steering command confirms that the control mechanism effectively translates perception outputs into appropriate steering actions for stable vehicle navigation. The system maintained stable performance under varying illumination conditions in the controlled indoor environment, indicating robustness of the perception and control pipeline. However, higher illumination levels occasionally reduced classification accuracy due to glare, reflections, and potential imbalance in the training dataset across different lighting conditions. Another factor affecting performance was the frame rate used for traffic sign classification. To optimize CPU and memory usage, the classification module processed two frames per second, which ensured stable system operation but may lead to missed detections.

The current implementation is limited to a structured indoor environment with predefined tracks and traffic signs. Real-world environments introduce additional challenges, including dynamic obstacles, complex road geometries, varying weather conditions, and diverse lighting scenarios, which may affect perception accuracy and control stability. Future work will focus on extending the system to more complex environments and improving the robustness of the perception and control modules.

\bibliographystyle{splncs04}
\bibliography{mybib}

\end{document}